# Novel Word Embedding and Translation-based Language Modeling for Extractive Speech Summarization

Kuan-Yu Chen, Shih-Hung Liu, Berlin Chen[+], Hsin-Min Wang, Hsin-Hsi Chen[#]
Academia Sinica, Taiwan
[#]National Taiwan University, Taiwan
[+]National Taiwan Normal University, Taiwan
{kychen, journey, whm}@iis.sinica.edu.tw, [+]berlin@csie.ntnu.edu.tw, [#]hhchen@csie.ntu.edu.tw

## ABSTRACT

Word embedding methods revolve around learning continuous distributed vector representations of words with neural networks, which can capture semantic and/or syntactic cues, and in turn be used to induce similarity measures among words, sentences and documents in context. Celebrated methods can be categorized as prediction-based and count-based methods according to the training objectives and model architectures. Their pros and cons have been extensively analyzed and evaluated in recent studies, but there is relatively less work continuing the line of research to develop an enhanced learning method that brings together the advantages of the two model families. In addition, the interpretation of the learned word representations still remains somewhat opaque. Motivated by the observations and considering the pressing need, this paper presents a novel method for learning the word representations, which not only inherits the advantages of classic word embedding methods but also offers a clearer and more rigorous interpretation of the learned word representations. Built upon the proposed word embedding method, we further formulate a translation-based language modeling framework for the extractive speech summarization task. A series of empirical evaluations demonstrate the effectiveness of the proposed word representation learning and language modeling techniques in extractive speech summarization.

## 1. INTRODUCTION

Owing to the popularity of various Internet applications, rapidly growing multimedia content, such as music video, broadcast news programs and lecture recordings, has been continuously filling our everyday life [1, 2]. Obviously, speech is one of the most important sources of information about multimedia. By virtue of speech summarization, one can efficiently browse multimedia content by digesting the summarized audio/video snippets and associated transcripts. Extractive speech summarization manages to select a set of salient sentences from a spoken document according to a target summarization ratio and subsequently concatenate them together to form a summary [3]. The wide spectrum of summarization methods developed so far may be roughly divided into three categories: 1) methods simply based on the sentence position or structure information [4], 2) methods based on unsupervised sentence ranking [5], and 3) methods based on supervised sentence classification [5]. Interested readers may refer to [5, 7, 8] for comprehensive reviews and new insights into the major methods that have been developed and applied with good success to a wide variety of text and speech summarization.

Orthogonal to the existing commonly-used methods, we explore in this paper the use of various word embedding methods [9-11] in extractive speech summarization, which have recently demonstrated excellent performance in many natural language processing (NLP) related tasks, such as machine translation [12], sentiment analysis [13] and sentence completion [14]. The central idea of these methods is to learn continuous, distributed vector representations of words using neural networks, which can probe latent semantic and/or syntactic cues, and in turn be employed to induce similarity measures among words, sentences and documents. According to the variety of the training objectives and model architectures, the classic methods can be roughly classified into the prediction-based and count-based methods [15]. Recent studies in the literature have evaluated these methods in several NLP-related tasks and analyzed their strengths and deficiencies [11, 16]. However, there are only a few studies in the literature that continue the line of research to crystalize an enhanced word embedding method that brings together the merits of these two major families. In addition, the interpretation of the learned value of each dimension in a learned word representation is a bit opaque. To satisfy the pressing need and complement the defect, we propose a novel modeling method, which not only inherits the advantages from the classic word embedding methods but also offers a clearer and more rigorous interpretation. Beyond the efforts to improve the representation of words, we also present a novel and efficient translation-based language modeling framework on top of the proposed word embedding method for extractive speech summarization. Unlike the common thread of leveraging word embedding methods in speech/text summarization tasks, which is to represent a document/sentence by averaging the corresponding word embeddings over all words in the document/sentence and estimate the cosine similarity measure of any given document-sentence pair, the proposed framework can authentically capture the finer-grained (i.e., word-to-word) semantic relationship to be effectively used in extractive speech summarization.

In a nutshell, the major contributions of the paper are twofold:

- **A novel word representation learning technique**, which not only inherits the advantages from the classic word embedding methods but also offers a clearer and more rigorous interpretation of word representations, is proposed.
- **A translation-based language modeling framework** on top of the proposed word embedding method, which can also be integrated with classic word embedding methods, is introduced to the extractive speech summarization task.

# 2. CLASSIC WORD EMBEDDING METHODS

Perhaps one of the most well-known seminal studies on developing word embedding methods was presented by Bengio et al. [9]. It estimated a statistical $n$-gram language model, formalized as a feed-forward neural network, for predicting future words in context while inducing word embeddings as a by-product. Such an attempt has already motivated many follow-up extensions to develop effective methods for probing latent semantic and syntactic regularities manifested in the representations of words. Representative methods can be categorized as prediction-based and count-based methods. The skip-gram model (SG) [10] and the global vector model (GloVe) [11] are well-studied examples of the two categories, respectively.

Rather than seeking to learn a statistical language model, the SG model is intended to obtain a dense vector representation of each word directly. The structure of SG is similar to a feed-forward neural network, with the exception that the non-linear hidden layer in the former is removed. The model thus can be trained on a large corpus efficiently, getting around the heavily computational burden incurred by the non-linear hidden layer, while still retaining good performance. Formally, given a word sequence, $w^1,w^2,\ldots,w^T$, the objective function of SG is to maximize the log-probability,

$$\sum_{t=1}^{T}\sum_{\substack{k=-c \\ k\neq 0}}^{c}\log P(w^{t+k}\mid w^t), \tag{1}$$

where $c$ is the window size of the contextual words for the central word $w^t$, and the conditional probability is computed by

$$P(w^{t+k}\mid w^t)=\frac{\exp(\mathbf{v}_{w^{t+k}}\cdot\mathbf{v}_{w^t})}{\sum_{i=1}^{V}\exp(\mathbf{v}_{w_i}\cdot\mathbf{v}_{w^t})}, \tag{2}$$

where $\mathbf{v}_{w^{t+k}}$ and $\mathbf{v}_{w^t}$ denote the representations of the words at positions $t+k$ and $t$, respectively; $w_i$ denotes the $i$-th word in the vocabulary; and $V$ is the vocabulary size.

The GloVe model suggests that an appropriate starting point for word representation learning should be associated with the ratios of co-occurrence probabilities rather than the prediction probabilities. More precisely, GloVe makes use of a weighted least squares regression, with the aim of learning word representations that can characterize the co-occurrence statistics between each pair of words:

$$\sum_{i=1}^{V}\sum_{j=1}^{V}f(X_{w_iw_j})(\mathbf{v}_{w_i}\cdot\mathbf{v}_{w_j}+b_{w_i}+b_{w_j}-\log X_{w_iw_j})^2, \tag{3}$$

where $w_i$ and $w_j$ are any two distinct words in the vocabulary; $X_{w_iw_j}$ denotes the number of times words $w_i$ and $w_j$ co-occur in a pre-defined sliding context window; $f(\cdot)$ is a monotonic smoothing function used to modulate the impact of each pair of words involved in model training; and $\mathbf{v}_{w_i}$ and $b_{w_i}$ denote the word representation and the bias term of word $w_i$, respectively. Interested readers may refer to [15, 16] for a more thorough and entertaining discussion.

# 3. METHODOLOGY

## 3.1 The Proposed Word Embedding Method

Although the prediction-based methods have shown their remarkable performance in several NLP-related tasks, they do not sufficiently utilize the statistics of the entire corpus since the models are usually trained on local context windows in a separate manner [11]. By contrast, the count-based methods leverage the holistic statistics of the corpus efficiently. However, a few studies have indicated their relatively poor performance in some tasks [16]. Among all the existing methods (both the prediction-based and count-based methods), the interpretation of the learned value of each dimension in the representation is not intuitively clear. Motivated by these observations, a novel modeling approach, which naturally brings together the advantages of the two major model families and results in interpretable word representations, is proposed.

We begin with the definition of terminologies and notations. As most classic embedding methods, we introduce two sets of word representations: one is the set of desired word representations, denoted by $\mathbf{M}$; the other is the set of separate context word representations, denoted by $\mathbf{W}$. $\mathbf{W}$ and $\mathbf{M}$ are $H\times V$ matrices, where the $j$-th columns of matrices $\mathbf{W}$ and $\mathbf{M}$, denoted by $\mathbf{W}_{w_j}\in\mathbb{R}^H$ and $\mathbf{M}_{w_j}\in\mathbb{R}^H$, correspond to the $j$-th word $w_j$ in the vocabulary. $H$ is a pre-defined dimension of the word embedding. To make the learned representation interpretable, we assume that each word embedding is a multinomial representation. Furthermore, to make the computation more efficient, we assume that each row vector of matrix $\mathbf{W}$ follows a multinomial distribution as well. To inherit the advantages from the prediction-based methods, the training objective is to obtain an appropriate word representation by considering the predictive ability of a given word occurring at an arbitrary position $t$ of the training corpus (denoted $w^t$) to predict its surrounding context words:

$$\prod_t\prod_{\substack{k=-c \\ k\neq 0}}^{c}P(w^{t+k}\mid w^t)=\prod_t\prod_{\substack{k=-c \\ k\neq 0}}^{c}\frac{\mathbf{W}_{w^{t+k}}\cdot\mathbf{M}_{w^t}}{\sum_{j=1}^{V}\mathbf{W}_{w_j}\cdot\mathbf{M}_{w^t}}. \tag{4}$$

The denominator can be omitted because it always equals to 1. In order to characterize the whole corpus statistics well, we train the model parameters in a batch mode instead of using a sequential learning strategy. Therefore, the objective function becomes

$$\sum_{i=1}^{V}\sum_{j=1}^{V}n(w_i,w_j)\log(\mathbf{W}_{w_i}\cdot\mathbf{M}_{w_j}), \tag{5}$$

where $n(w_i,w_j)$ denotes the number of times words $w_i$ and $w_j$ co-occur in a pre-defined sliding context window. Obviously, such a model not only bears a close resemblance to the prediction-based methods (e.g., SG) but also capitalizes on the statistics gathered from the entire corpus like the count-based methods (e.g., GloVe), in a probabilistic framework. The component distributions (i.e., $\mathbf{W}$ and $\mathbf{M}$) can be estimated using the expectation-maximization (EM) algorithm. Advanced algorithms, such as the triple jump EM algorithm [17], can be leveraged to accelerate the training process. Since the training objective of the proposed method is similar to that of the SG model, and it results a set of distributional word representations, we thus term the proposed model the distributional skip-gram model (DSG). The interpretive ability of DSG will be discussed in detail later in Section 4.3.

## 3.2 Translation-based Language Modeling for Summarization

Language modeling (LM) has proven its broad utility in many NLP-related tasks. In the context of using LM for extractive speech summarization, each sentence $S$ of a spoken document $D$ to be summarized can be formulated as a probabilistic generative model for generating the document, and sentences are selected based on the corresponding generative probability $P(D|S)$: the higher the probability, the more representative $S$ is likely to be for $D$ [18]. The major challenge facing the LM-based framework is how to accurately estimate the model parameters for each sentence. The simplest way is to estimate a unigram language model (ULM) on the basis of the frequency of each distinct word $w$ occurring in the sentence $S$, with the maximum likelihood criterion:

$$P(w \mid S) = \frac{n(w,S)}{|S|}, \quad (6)$$

where $n(w,S)$ is the number of times that word $w$ occurs in sentence $S$, and $|S|$ is the length of the sentence.

There is a general consensus that merely matching terms in a candidate sentence and the document to be summarized may not always select summary sentences that can capture the important semantic intent of the document. Thus, in order to more precisely assess the representativeness of a sentence to the document, we suggest inferring the probability that the document would be generated as a translation of the sentence. That is, the generating probability is calculated based on a translation model of the form $P(w_j|w_i)$, which is the probability that a sentence word $w_i$ is semantically translated to a document word $w_j$:

$$P(D \mid S) = \prod_{w_j \in D} \left[ \sum_{w_i \in S} P(w_j \mid w_i) P(w_i \mid S) \right]^{n(w_j,D)}. \quad (7)$$

Accordingly, the translation-based language modeling approach allows us to score a sentence by computing the degree of match between a sentence word and the semantically related words in the document. If $P(w_j|w_i)$ only allows a word to be translated into itself, Eq. (7) will be reduced to the ULM approach (cf. Eq. (6)). However, $P(w_j|w_i)$ would in general allow us to translate $w_i$ into the semantically related words with non-zero probabilities, thereby achieving "semantic matching" between the document and its component sentences.

Based on the proposed DSG method, the translation probability $P(w_j|w_i)$ can be naturally computed by:

$$P(w_j \mid w_i) = \mathbf{W}_{w_j} \cdot \mathbf{M}_{w_i}. \quad (8)$$

Consequently, the sentences offering the highest generated probability (cf. Eqs (7) and (8)) and dissimilar to those already selected sentences (for an already selected sentence $S'$, computing $P(S|S')$ using Eq. (7)) will be selected and sequenced to form the final summary according to a desired summarization ratio. The proposed translation-based language modeling method is denoted by TBLM hereafter.

# 4. EXPERIMENTS

## 4.1 Dataset & Setup

We conduct a series of experiments on a Mandarin Benchmark broadcast new corpus [19]. The MATBN dataset is publicly available and has been widely used to evaluate several NLP-related tasks, including speech recognition [20], information retrieval [21] and summarization [18]. As such, we follow the experimental setting used by some previous studies for speech summarization. The average word error rate of the automatic transcripts of these broadcast news documents is about 38%. The reference summaries were generated by ranking the sentences in the manual transcript of a broadcast news document by importance without assigning a score to each sentence. Each document has three reference summaries annotated by three subjects. For the assessment of summarization performance, we adopted the widely-used ROUGE metrics (in F-scores) [22]. The summarization ratio was set to 10%. In addition, a corpus of 14,000 text news documents, compiled during the same period as the broadcast news documents, was used to estimate the parameters of the models compared in this paper.

## 4.2 Experimental Results

A common thread of leveraging word embedding methods in a summarization task is to represent a document/sentence by averaging the corresponding word embeddings over all words in the document/sentence. After that, the cosine similarity measure, as a straightforward choice, can be readily applied to determine the degree of relevance between any pair of representations [29, 30]. In the first place, we try to investigate the effectiveness of two state-of-the-art word embedding methods (i.e., SG and GloVe) and the proposed method (i.e., DSG), in conjunction with the cosine similarity measure for speech summarization. The experimental results are shown in Table 1, where MT denotes the results obtained based on the manual transcripts of spoken documents, and SRT denotes the results using the speech recognition transcripts of spoken documents that may contain speech recognition errors. Several observations can be made from these results. First, the two classic word embedding methods, though based on disparate model structures and learning strategies, achieve results competitive to each other in both the MT and SRT cases. Second, the proposed DSG method, which naturally brings together the advantages of the two major model families (i.e., prediction-based and count-based) in the literature, outperforms SG and GloVe (representatives of the two model families, respectively) by a significant margin in both the MT and SRT cases. Third, since the relevance degree between a document-sentence pair is computed by the cosine similarity measure, vector space-based methods, such as VSM [23], LSA [23] and MMR [24], can be treated as the principled baseline systems. Albeit that the classic word embedding methods (i.e., GloVe and SG) outperform the conventional VSM model, they achieve almost the same level of performance as LSA and MMR, which are considered two enhanced versions of VSM. It should be noted that the proposed DSG method not only outperforms VSM, but also is superior to LSA and MMR in both the MT and SRT cases.

**Table 1. Summarization results achieved by various word-embedding methods in conjunction with the cosine similarity measure.**

| Cosine | MT | | SRT | |
|---|---|---|---|---|
| | ROUGE-2 | ROUGE-L | ROUGE-2 | ROUGE-L |
| SG | 0.239 | 0.311 | 0.215 | 0.311 |
| GloVe | 0.244 | 0.310 | 0.214 | 0.310 |
| DSG | **0.281** | **0.351** | **0.234** | **0.330** |
| VSM | 0.228 | 0.290 | 0.189 | 0.287 |
| LSA | 0.233 | 0.316 | 0.201 | 0.301 |
| MMR | 0.248 | 0.322 | 0.215 | 0.315 |

**Table 2. Summarization results achieved by various word-embedding methods in conjunction with the translation-based language modeling method.**

| TBLM | MT | | SRT | |
|---|---|---|---|---|
| | ROUGE-2 | ROUGE-L | ROUGE-2 | ROUGE-L |
| SG | 0.320 | 0.385 | 0.225 | 0.322 |
| GloVe | 0.309 | 0.372 | 0.239 | **0.332** |
| DSG | **0.333** | **0.389** | **0.244** | 0.331 |
| ULM | 0.298 | 0.362 | 0.210 | 0.307 |

**Table 3. Summarization results achieved by a few well-studied or/and state-of-the-art unsupervised methods.**

| | MT | | SRT | |
|---|---|---|---|---|
| | ROUGE-2 | ROUGE-L | ROUGE-2 | ROUGE-L |
| MRW | 0.282 | 0.332 | 0.191 | 0.291 |
| LexRank | 0.309 | 0.305 | 0.146 | 0.254 |
| SM | 0.286 | 0.332 | 0.204 | 0.303 |
| ILP | **0.337** | 0.348 | 0.209 | 0.306 |
| DSG(TBLM) | 0.333 | **0.389** | **0.244** | **0.331** |

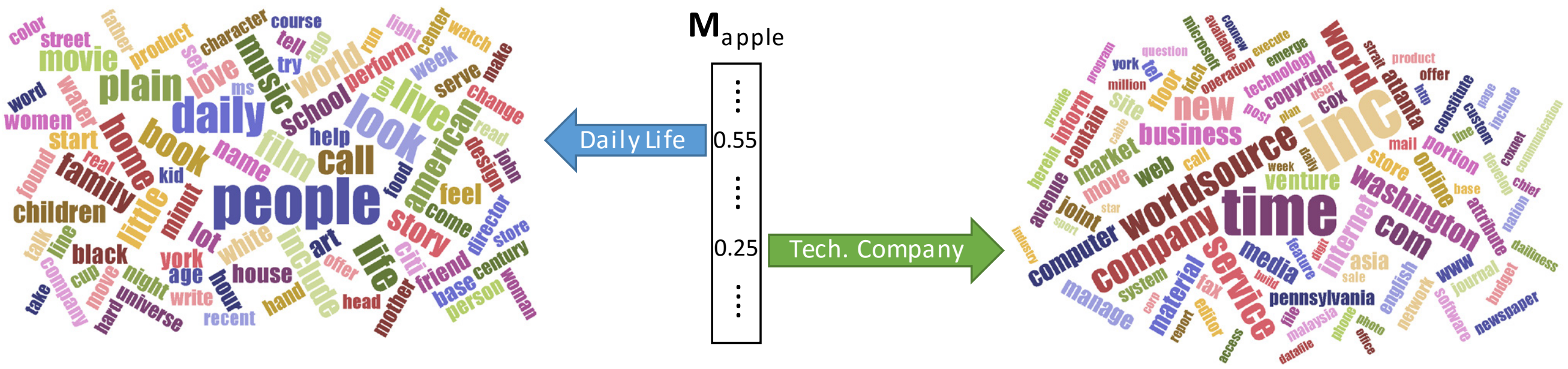

**Figure 1. A running example for interpreting the word embeddings learned by DSG.**

Next, we evaluate the proposed TBLM method. It is worth mentioning that TBLM is well suitable for pairing with the DSG method since the translation probability can be easily obtained by lightweight calculation (cf. Eq. (8)). It can also be integrated with the classic word embedding methods (e.g. SG and GloVe), but with a heavier computational burden (cf. Eq. (2) for example). The experimental results are summarized in Table 2. Three observations can be made from the results. First, since the proposed DSG method inherits the advantages from prediction- and count-based methods, it outperforms both SG and GloVe, when all the three models are paired with the proposed TBLM method. Second, when integrated with TBLM, all the three word embedding methods outperform the baseline ULM method [18, 31] (cf. Section 3.2) by a remarkable margin in both the MT and SRT cases. Third, comparing the results in Tables 1 and 2, it is evident that TBLM is deemed a preferable vehicle to make use of powerful word embedding methods in speech summarization.

In the last set of experiments, we assess the performance levels of several well-practiced unsupervised summarization methods, including the graph-based methods (i.e., MRW [25] and LexRank [26]), the submodularity (SM) method [27], and the integer linear programming (ILP) method [28]. The corresponding results are illustrated in Table 3. The performance trends of these state-of-the-art methods in our study are quite in line with those observations made by other previous studies in different extractive summarization tasks. A noticeable observation is that speech recognition errors may lead to inaccurate similarity measures between a pair of sentences or document-sentence. Probably due to this reason, the two graph-based methods (i.e., MRW and LexRank) cannot perform on par with the vector-space methods (i.e., VSM, LSA, and MMR) (cf. Table 1) in the SRT case, but the situation is reversed in the MT case. Moreover, the SM and ILP achieve the best results in the MT case, but only offer mediocre performance among all methods in the SRT case. To sum up, the proposed DSG method, which inherits the advantages from both the prediction- and count-based methods, indeed outperforms classic word embedding methods when paired with different summarization strategies (i.e., the cosine similarity measure and TBLM). The proposed TBLM method further enhances the DSG method since it can authentically capture a finer-grained (i.e., word-to-word) semantic relationship to be effectively used in extractive speech summarization.

## 4.3 Further Analysis

SG, GloVe and the proposed DSG method can be analyzed from several critical perspectives. First, SG and DSG aim at maximizing the collection likelihood in training, while GloVe concentrates on discovering useful information from the co-occurrence statistics between each pair of words. It is worthy to note that GloVe has a close relation with the classic weighted matrix factorization approach, while the major difference is that the former concentrates on rendering the *word-by-word* co-occurrence matrix while the latter decomposes the *word-by-document* matrix [23, 32, 33]. Second, since the parameters (i.e., word representations) of SG are trained sequentially (i.e., the so-called *on-line* learning strategy), the sequential order of the training corpus may make the resulting models unstable. On the contrary, GloVe and DSG accumulate the statistics over the entire training corpus in the first place; the model parameters are then updated based on such censuses at once (i.e., the so-called *batch-mode* learning strategy). Finally, the word vectors learned by DSG are distributional representations, while SG and GloVe express each word by a distributed representation.

The classic embedding methods usually output two sets of word representations, but there is no particular use for the context word representations. Since DSG assumes that each row of the context word matrix **W** follows a multinomial distribution (i.e., a multinomial distribution over words), we can naturally interpret the semantic meaning of each dimension of a word embedding by referring to the words with higher probabilities in the corresponding row vector in **W**. Since each word embedding in the desired word representations **M** is also a multinomial distribution, we can interpret a learned word embedding by first identifying the dimensions with higher probabilities and then identifying the context words with higher probabilities in the corresponding row vectors of **W**. Figure 1 shows a running example for word "apple" learned by DSG on the LDC Gigaword corpus. The word cloud can be plotted with respect to the probabilities of individual context words for a selected dimension. It is obvious that "apple" is not only a kind of important matter in our daily life, but also is a famous technology company. The example shows that the word embeddings learned by DSG can be interpreted in a reasonable and systematical manner.

# 5. CONCLUSIONS

A novel distributional word embedding method and a translation-based language modeling method have been proposed and introduced for extractive speech summarization in this paper. Empirical results have demonstrated their respective and joint effectiveness and efficiency over several state-of-the-art summarization methods. In the future, we plan to further extend and apply the proposed framework to a wider range of summarization and NLP-related tasks. We will also concentrate on integrating a variety of prior knowledge for learning the word representations.